# Smart Textile-Driven Soft Spine Exosuit for Lifting Tasks in Industrial Applications

Kefan Zhu, Bibhu Sharma, Phuoc Thien Phan, James Davies, Mai Thanh Thai, Trung Thien Hoang, Chi Cong Nguyen, Adrienne Ji, Emanuele Nicotra, Nigel H. Lovell, and Thanh Nho Do*

*Abstract*—Work-related musculoskeletal disorders (WMSDs) are often caused by repetitive lifting, making them a significant concern in occupational health. Although wearable assist devices have become the norm for mitigating the risk of back pain, most spinal assist devices still possess a partially rigid structure that impacts the user's comfort and flexibility. This paper addresses this issue by presenting a smart textile-actuated spine assistance robotic exosuit (SARE), which can conform to the back seamlessly without impeding the user's movement and is incredibly lightweight. The SARE can assist the human erector spinae to complete any action with virtually infinite degrees of freedom. To detect the strain on the spine and to control the smart textile automatically, a soft knitting sensor which utilizes fluid pressure as sensing element is used. The new device is validated experimentally with human subjects where it reduces peak electromyography (EMG) signals of lumbar erector spinae by 32%±15% in loaded and 22%±8.2% in unloaded conditions. Moreover, the integrated EMG decreased by 24.2%±13.6% under loaded condition and 23.6%±9% under unloaded condition. In summary, the artificial muscle wearable device represents an anatomical solution to reduce the risk of muscle strain, metabolic energy cost and back pain associated with repetitive lifting tasks.

## I. Introduction

Work-related musculoskeletal disorders (WMSDs) are still jeopardizing the health of workers. Within WMSDs, approximately 40% are estimated to be back-related [1]. Prolonged standing and repetitive handling have been established as significant risk factors for WMSDs. For example, in aviation, airport baggage handlers have lifting, carrying and holding tasks, which are known to induce lumbar disc herniation and lower back pain (LBP) [2]. Studies have shown that 62.7% of handlers have LBP [3]. Further, LBP incidence in baggage handlers increases significantly with time [4]. While squat lifting is preferable over stoop lifting, as it hinders forward flexion of the spine, reducing the internal spinal loading [5], stoop lifting cannot be completely avoided. This means preventing WMSDs associated with stoop lifting should receive significant priority.

This work was supported by a UNSW Scientia Fellowship.

All authors are with the Graduate School of Biomedical Engineering, UNSW Sydney, NSW, 2052, Australia. (email: kefan.zhu@student.unsw.edu.au; bibhu.sharma@unsw.edu.au; phuoc_thien.phan@unsw.edu.au; j.j.davies@student.unsw.edu.au; trungthien.hoang@unsw.edu.au; maithanh.thai@unsw.edu.au; cong.c.nguyen@unsw.edu.au; e.nicotra@unsw.edu.au; n.lovell@unsw.edu.au; tn.do@unsw.edu.au

M.T. Thai is with the College of Engineering and Computer Science, VinUniversity, Hanoi, Viet Nam.

[1]Corresponding author, email: tn.do@unsw.edu.au

Numerous active wearable devices have been proposed. Pelvic joint assistive devices apply an additional moment on the pelvic joint rotation in the sagittal plane [6]. A device developed in [7] utilizes two motors placed on either side of the hip joint to generate torque on the joint, thereby assisting in trunk flexion and extension. It could reduce approximately 34.0% the average integrated electromyogram (iEMG) in erector spinae during repetitive lifting. Another representative pelvic joint assistive device is hybrid assistive limb (HAL) lumbar support, which reduced 14% of myoelectric activity of lumbar erector muscles [8]. Compared to HAL, a lower-back robotic exoskeleton exhibited a reduction of approximately 34% in the erector spine muscles [9]. However, such rigid exoskeletons typically exhibit considerable weight, thereby introducing additional mass and inertia at the distal segments of the user's body. These factors can burden the metabolic cost of the user's movements [10]. Furthermore, the misalignment of joints is one inevitable problem evident with rigid wearable devices, which may compromise the original physiological exercise by producing unsatisfactory interaction forces on the human limbs [10].

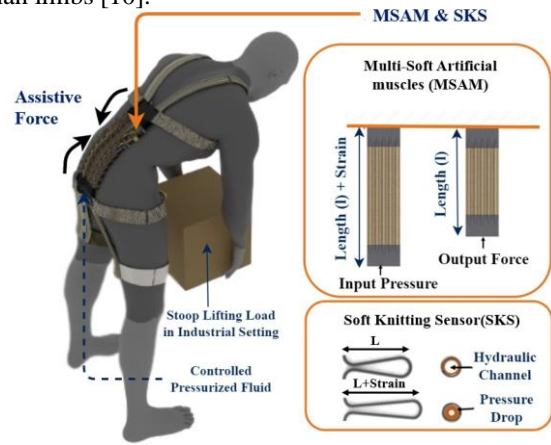

Figure 1 Overview of the spine assistive system for stoop lifting task based on Multi-Soft Artificial muscles (MSAM) and soft knitting sensor (SKS).

Spine assistive devices apply a compression force, which acts in parallel with back muscles such as the lumbar erector spinae (LES) or latissimus dorsi (LD) to assist their function [6]. For example, a cable-driven soft spine wearable device consisting of a hyper-redundant continuum mechanism was developed to mimic the compression force of ES muscles, which was observed to reduce LES usage by 30% [11]. Another soft power suit utilizes two twisted string actuators (TSA) fixed at the lower back, reducing by 21.4-25.2% the muscle activation required for dynamic lifting [12]. However, these actuators are associated with high friction loss, low speed, and high nonlinear hysteresis, making them challenging to control. They also contain rigid structures, which do not interface well with the human body, reducing the comfort and portability of the devices [13].

Regarding passive devices, they are found to be better suited for light assistance tasks with limited dynamic movements. An analysis involving a passive personal lift assistive device demonstrated that the decrement in metabolic cost for lifting was not significant [14]. Moreover, as stoop lifting is highly dynamic and demands considerable joint load, it would necessitate the use of more sophisticated active solutions [15]. To address the aforementioned issues through active assistance, this paper will develop a soft and lightweight spine assistive robotic exosuit which can provide useful assistance during repetitive lifting tasks with an aim to reduce the risk of back-related musculoskeletal disorders (*Fig. 1*).

## II. MATERIALS & METHODS

### A. Mechanical Design

#### 1) Device overview and working principle

The SARE is comprised of two essential components integrated into a smart textile - the Soft Knitting Sensor (SKS) and the Multi-Soft Artificial Muscles (MSAM). The SKS is a highly conformable sensor with a knitting technique that uses a hydraulic pressure-based mechanism to detect strain and curvature of the spine. When attached to the human back, the SKS adjusts with the curvature of the spine during lifting activities. As the curvature increases, the inner hydraulic pressure decreases, and this change is detected by the SKS, thereby informing about the spinal curvature in real-time.

The MSAM, on the other hand, acts as the active component of the SARE. Constructed from soft microtubules (NuSil™ Technology LLC, Carpinteria, CA, USA) and helical coils (McMaster-Carr Supply Co., Elmhurst, IL, USA), based on hydraulic filament artificial muscles [16-20]. A plastic guide tube is connected between a Luer-lock™ 3-mL syringe (BD, USA) and the artificial muscle for transmitting hydraulic pressure to the artificial muscle. The working principles of the MSAM are as follows: in its resting state, the MSAM is elongated due to the pressurization achieved by pushing the syringe. For generating compressive force, releasing the piston of syringe, the stored potential energy within the microtubules and helical coils triggers the contraction of the MSAM, thereby lifting the load (*Fig. 2c*). It is mimicking the function of the human erector spinae muscles.

There are five fabrication process steps utilize MSAM (*Fig. 2a*). First step, preparing two layers of axially stretchable fabrics and fit together. This stretchable fabric is chosen as a commercial self-adhesive compression bandage (3M, Australia), which is Lightweight, porous, and comfortable for patients. The metal structure of MSAM is fully covered inside the fabric without contacting human skin. Step 2, hollow channels are made for inserting artificial muscles for combining artificial muscles with the compression bandage by Zigzag stitch. The hollow channels offer a distinct advantage by effectively limiting the radial strain arising from the torsional forces exerted by the helical coil. The boundary of the hollow channel is created with a stitching line produced by a sewing machine. A metal rod is inserted between two stitched lines to expand the hollow space (*Fig. 2b*). This facilitates the easier passage of the artificial muscles through the channels. Step 3, after hollow channels are completed, each artificial muscle is individually threaded through its respective hollow channels and aligned in parallel. Step 4, then non-stretchable fabric is sewed to the two ends of the stretchable fabric to fixate the artificial muscles. Step 5, linking the MSAM to wearable belts, which are made of comfortable nylon material. The upper portion is wrapped around the chest and passes over the shoulders to ensure secure fastening. The lower portion is wrapped around the pelvic and thigh regions (*Fig. 2d*). Belts are adaptable to user dimensions by adjustable buckles. During the lifting in a stooped position, the MSAM extend parallel to the spine while the SARE pivots firmly on the distal and proximal supports. The fabrication process is illustrated in *Fig. 2a*.

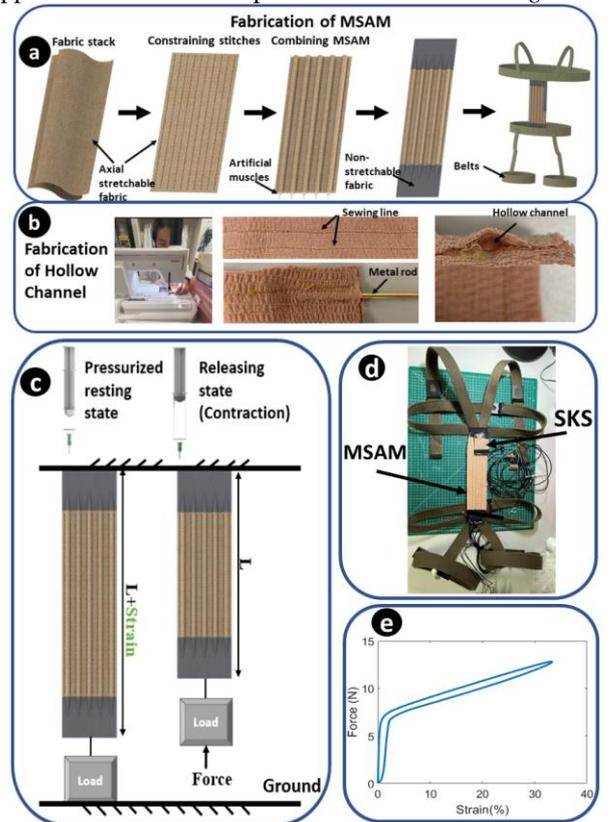

Figure 2 Fabrication and prototype. a) Overview of fabrication of MSAM. b) Steps for fabricating hollow channel with sewing machine and metal rob. c)Working principle of MSAM. d) Prototype of SARE. e) The compression force-strain profile for individual artificial muscle.

#### 2) Mathematical modelling of the musculoskeletal system

Taking inspiration from the force transmission within the muscle of the human back informs the development of an active fabric structure that is both comfortable and efficient in design [12]. The ES is an essential structure for maintaining the upright posture of the human body, which includes the iliocostalis, longissimus, and spinalis. It is responsible for the extension of the trunk and stabilizing the lumbar spine. From a thoracolumbar finite element (FE) Model for a lumbar posture, the maximum axial strain of the spine was estimated at 20% during stoop lifting tasks [21]. Also, the total predicted force on the LES was 242 N for 45º flexion bending of the hip [22]. The mathematical modelling of the musculoskeletal system during lifting tasks can be intricate due to the numerous joints and muscles involved. While the FE method can provide numerical solutions [21], achieving convergence can be challenging due to the multitude of variables involved. However, a linked-segment model, as [23] can be used to simplify the human limb to minimize the degrees of freedom (DOF). For the stoop lifting task, this paper considers a 2-DOF linked model, which has two joints: hip and L5-S1 joint. The angular displacement of lumbar-sacral joint represents lordosis/kyphosis of the spine. Consider the generalized co-ordinate $\theta \in \mathbb{R}^2$ as shown in *Fig. 3*, actuated by joint torques $\tau \in \mathbb{R}^2$. When lifting a load of mass ($m_{load}$), the moment created at the elbow joint is $\tau_L$. The assistance provided by the spine assistance device can be inferred as a torque $\tau_{ext}$. The Lagrangian formulation, which involves the evaluation of kinetic function and potential function, was used to construct a canonical relation between kinetic and kinematic parameters as:

$$M(\theta)\ddot{\theta} + C(\theta,\dot{\theta})\dot{\theta} + G(\theta) = \tau + J(\theta)^T \tau_L - J_2(\theta)^T \tau_{ext} \quad (1)$$

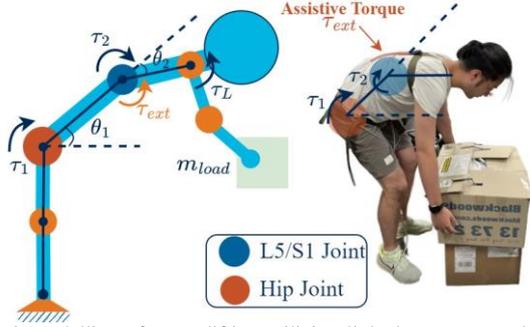

Figure 3 Modelling of stoop-lifting utilizing linked segment model of the human musculoskeletal system.

Here, the Jacobian $J(\theta)$ describes the relationship between end-effector forces and joint torques and $J_2(\theta)$ is the Jacobian that describes the relation between assistive force $\tau_{ext}$ and joint torques. While $J(\theta)$ is derived using the velocity relation between end-effector and joint, $J2(\theta)$ is simply $J_2(\theta) = [0\ 1]$, as assistive force is assumed to apply only on the L5-S1 joint. $M(\theta)$ and $C(\theta,\dot{\theta})$ are mass matrix and coriolis matrix, which depend upon the biomechanical parameters of the human subject. $G(\theta)$ is the gravity component. These equations are essential for determining the kinetic parameters that inform the design process and are also critical for the development of functional control systems. The validation of the model was conducted by comparing experimental data from [24] (*Fig. 4*) and the biomechanical parameters required in (1) also derived from [24].

*3) Actuation Mechanism*

Based on the anatomy and developed mathematical model, an actuation mechanism was designed. The composite of MSAM is comprising 126 N/m spring constant helical coils and soft microtubules (E=1.2556 MPa) and related steel tubes, guide tubes, and syringes. In order to provide compression force that reduces 20% LES force, which is 50 N. According to the output force equation generated by the artificial muscle is determined in [16] as:

$$F_{out} = \alpha E A_t \left(1 - \frac{1}{1+\frac{x}{l_i}}\right) + k_c x \quad (2)$$

where $\alpha$ is the pre-sketched strain of the microtubule for inserting into the helical coil, $E$ is Young's Modulus for the soft microtubules, $A_t$ is the cross-sectional area of the soft microtubules, $k_c$ is the spring constant of the helical coil, $x$ is the elongation value and $l_i$ is the initial length of the artificial muscle. MASM is designed to contain five artificial muscles. For 20% MSAM strain, 2 ml volume of liquid from 3 ml standard syringes (Livingstone International Pty. Limited, AUS) with 55.4 mm$^2$ piston area are used. The thickness of each MSAM layer is 5 mm, which can be concealed underneath the garments completely. The MSAM weight was 56.8 g and can apply 35 N over the entire 60 mm stroke range and a maximum of 50 N at 360 mm length (*Fig. 2e*).

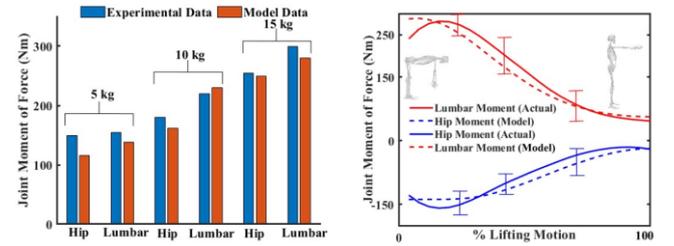

Figure 4 Comparison of developed mathematical model with experimental data of stoop lifting from [20]. Left figure shows peak hip torque and lumbar torque during stoop lifting of various weights. Right figure compares hip torque and lumbar torque between actual and model data.

To accurately control the strain of the MSAM, a linear actuator was employed to the piston of the syringe. The stroke of the linear actuator was controlled to alter the hydraulic pressure, thus realizing the desired strain of the artificial muscle. To enhance portability and comfort, the actuator and wearable components were separated. In order to determine an appropriate actuator, it is imperative to meet the theoretical requirements for thrust. The calculation of thrust involves both the internal pressure of the syringe and the pressure area of the piston.

Using (3) [16], the internal pressure was determined to be 3.81 MPa, leading to a requirement of 1055.2 N of thrust to support MSAM. As a result, the LACT8-1000BPL Linear actuator (Concentric™, Des Moines, Iowa, USA) was selected.

$$P = F_{out} \left(\frac{\pi}{4} d_o^2 - \frac{\alpha A_t}{1+x/l_i}\right)^{-1} \quad (3)$$

where $d_o$ is the outer diameter of the microtubules.

*4) Soft Knitting Sensor*

Same as MSAM, the development of SKS is achieved through the integration of helical coil, soft microtubule and extra pressure sensor. The soft knitting sensor (SKS) presented in this study draws inspiration from the hydraulic soft filament sensor and employs a knitting technique, which is capable of detecting changes in inner hydraulic pressure due to strain by integrated pressure sensor, informing the curvature of the spine from the measurement of hydraulic pressure. *Fig. 5c* illustrates this relationship, displaying the working principle and patterns that emerge as the SKS reacts to variations in spinal curvature. The SKS is attached to the MSAM. Because MSAM shares curvature and strain with the spine. it is stretched with the curvature of the MSAM during stoop lifting, causing the inner hydraulic pressure to drop with increased curvature (*Fig. 5e*) and pressure changed of One-time stoop lifting, based on strain of SKS (*Fig. 6a*). By examining the relationship between pressure and strain of SKS, curvature of the spine can be determined from inner pressure signal.

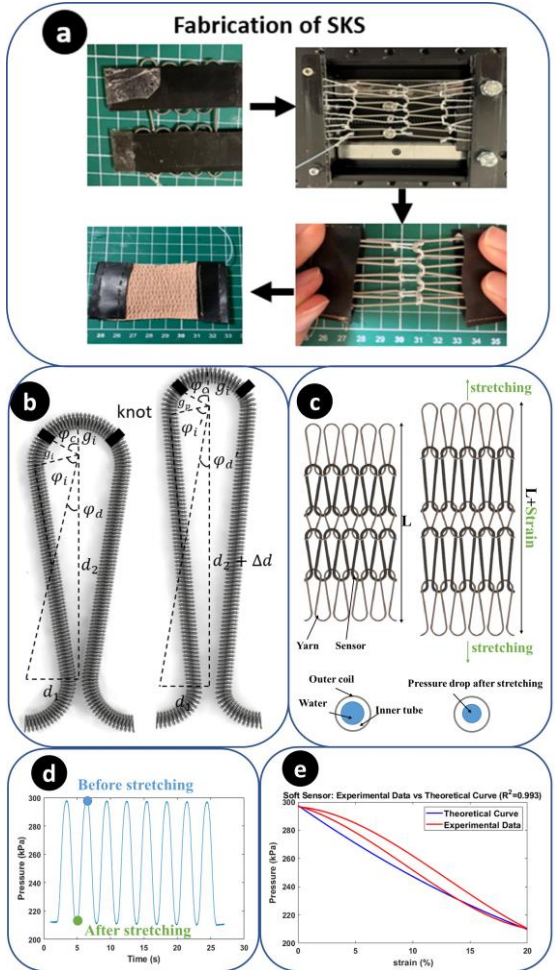

Figure 5 Sensing system along with the control system. a) Overview of the SKS. b) Left side is the parameter distribution of mathematical models. c) working principle of SKS d) The pressure signal of the SKS is changing during an 8 turns sine wave stretching. e) The pressure-strain profile of the SKS under experimental data and theoretical data ($R^2$=0.993).

The knitting technique offers significant advantages in high conformability and an interconnected structure. The knitted fabrics allow them to conform making them well-suited for applications involving direct contact with the human body. The interconnected structure allows for enhanced sensitivity by dividing the sensor into multiple short elements, enabling precise interpretation of subtle changes in deformation or pressure. The fabrication of SKS is shown in *Fig. 5a*. For the knitting technique, the sensor is bent to fit into a pre-designed pattern of two knitting tools. Subsequently, each loop is knotted using the same loop-shaped yarn, and the other side of the yarn is connected to another layer loop. Consequently, the sensor can be stretched by pulling on the two ends of the yarn, like *Fig. 5c*. Finally, an axial stretchable fabric is sewed by sewing machine, to the surface of the sensors to limit the radial strain and a non-stretchable fabric was combined with two ends of yarn to fixate SKS. In order to prevent stitching from inhibiting vertical elongation, the fabric is stretched during sewing. The initial inner pressure of SKS is 300 kPa, with a drop-in pressure of 90 kPa as elongation reaches 20% strain. The SKS demonstrates a sensitivity of 4.50 kPa/%, and the threshold is set at 280 kPa to prevent inadvertent activation of the linear actuator. For the test, SKS is fixed in a platform of linear actuator. The linear actuator provides a sin wave movement function to elongate SKS. The pressure change results are shown in *Fig. 5d*.

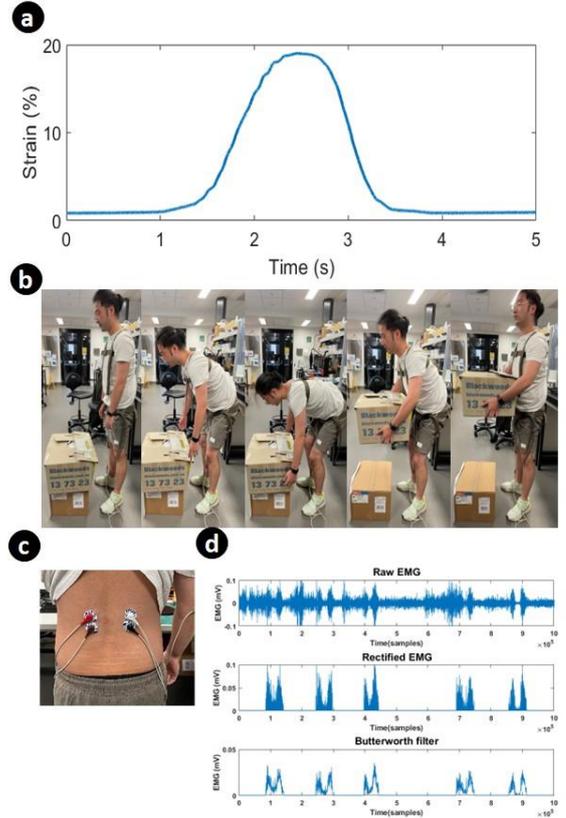

Figure 6 Outline of the experimental setup. a) The pressure signal vs one-time lifting is converted to strain signal by the pressure-strain profile. b) Demonstration of stoop lifting with SARE, and the posture corresponding to the strain signal with time. c) Demonstration of electrodes position for LES EMG experiments. d) Data Processing of EMG from raw data to filtered data.

To develop a mathematical model for the SKS, certain assumptions are made. Firstly, the silicone microtubule is considered to be a linearly elastic material under conditions of 300 kPa pressure and 20% strain. Secondly, the vertical distance between two centres of semicircle ($d_2$) is increasing linearly. Thirdly, the helical coil used in the sensor constrains the radial expansion of the microtubule. The SKS provides measurements of four parameters: the half horizontal distance between centres of two loops ($d_1$), the half vertical distance between centres of two loops ($d_2$), the radius of semicircle ($g_i$) and the angle between knot position and vertical vector ($\varphi_c$). After the sensor is stretched, two main variables that need consideration are the changed radius of semicircle ($g_p$) and the difference in vertical distance between two centres of a semicircle after stretching ($\Delta d$). Geometrically, $g_p$ and $\Delta d$ both can be determined by following equations with respect to strain.

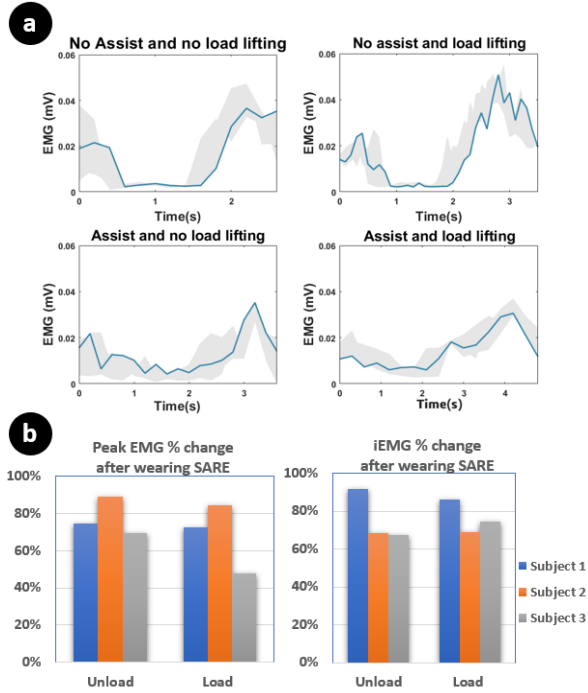

Figure 7 Results of processed EMG signal for four experiments. a) Overview of the entire lifting period with a combination of all valid tasks for different lifting types. b) Column plot of Peak EMG signal and iEMG signal percentage change for three subjects.

$$\Delta d = strain \times (2g_i \times cos(\varphi_c) + d_2) \quad (4)$$

$$\Delta d = 2\sqrt{(strain+1)^2 \times \left(\frac{d_1^2 + d_2^2}{4} - g_i^2\right) - \frac{d_1^2 + d_2^2}{4} + g_p^2} \quad (5)$$

Based on Barlow's formula, $P_B = \frac{2TS}{D}$ (6)

Where $P_B$ is the inner pressure of tube, $T$ is the thickness of tube wall, $S$ is the allowable stress of tube and $D$ is the outside diameter of tube. The relationship between initial inner pressure and changed inner pressure can be determined by (6). Then, combined with (4) and (5), the inner pressure change of SKS ($P'$) can be defined with one variable strain.

$$\frac{P}{P'} = \frac{\pi \times g_p \times \frac{180 - \varphi_d' - \varphi_c - \varphi_i'}{180} + \sqrt{\frac{d_1^2 + (d_2 + \Delta d)^2}{4} - g_p^2} + \frac{2g_i \times \pi \times \varphi_c}{180}}{\pi \times g_i \times \frac{180 - \varphi_d - \varphi_c - \varphi_i}{180} + \sqrt{\frac{d_1^2 + d_2^2}{4} - g_i^2} + \frac{2g_i \times \pi \times \varphi_c}{180}} \quad (7)$$

Where $\varphi_i, \varphi_d, \varphi_i', \varphi_d'$ are the angles in the geometrical model for calculation (*Fig. 5b*). P is the initial inner pressure of the sensor. The pressure-strain profile of the SKS under experimental data and theoretical data are shown in *Fig. 5e*. The observed cubic curve characteristic in the experimental data can be seen as a combination of delay of SKS and quadratic curve (Theoretical data). The delay of SKS could be attributed to the pre-stretching of the silicone tube, slight slippage of knots with yarn during the fabrication process, or the sensor may not have been fully stretched initially at 0% strain.

*5) Surface electromyography (sEMG)*

EMG aids in the detection and monitoring of the electrical activity of muscles. sEMG uses surface electrodes to detect multiple motor units above the skin. It provides a quantitative method to assess muscle function, movement patterns, and localized muscle fatigue to inform clinical decision-making processes. A PowerLab 26T (AD Instruments, Sydney), which collects two signals simultaneously at a 10 kHz sampling rate, is used for data acquisition. Pre-gelled conductive adhesive recording electrodes are used on the skin surface. The position of electrodes is placed 5 cm laterally from the median line of the third lumbar vertebra for detecting LES signal. Ground electrodes are placed on the distal upper arm (Fig. 6c).

6) Data Collection

The male volunteers (N =3, height = 179 cm±3.7 and weight = 67.3 kg±8.7) of this experiment were recruited based on the absence of a history of back pain and being in good health. The experiment was conducted under the UNSW Ethics Approval No. HC210717. Before the experiment, the volunteers were instructed to relax their spines for 30 minutes.

All four experiments were completed by stoop lifting, whereby the volunteers were required to maintain a straight lower lumbar and flex at the hip joint. A 5 kg weight positioned. 5 cm in front of the volunteer's toe. Upon touching the bottom of the weight after stooping, the volunteers return to an erect standing posture, thereby maintaining a consistent bending angle of the back. The data were collected for four scenarios. In the first scenario, the subject mimicked stoop lifting without any load. The second scenario involved the subject carrying a load, and both scenarios were initially performed without any assistance. Subsequently, the experiments were repeated with the assistance of SARE, and the EMG data was recorded from all four scenarios for further analysis (*Fig. 6b).*

*7) Data Processing*

EMG signal processing was divided into four parts: bandpass filter, rectification, smoothing and integration. Bandpass filter (50-400 Hz) was applied through the digital processing in PowerLab 26T data acquisition system. Smoothing was achieved by MATLAB with Butterworth function (10 Hz), designed to reduce shape of frequency response. *Fig. 6d* shows the rectified and smoothing process steps of a raw EMG signal recorded from the LES. After filtering the EMG signal, the iEMG signal was calculated by integrating the area under the curve of the rectified EMG signal. The efficiency of the assist device is calculated as:

$$E = 1 - \frac{EMG_{assist}}{EMG_{no-assist}} \quad (9)$$

$EMG_{assist}$ and $EMG_{no-assist}$ are the iEMG signal when the volunteer tests with/without SARE.

III. RESULT

From *Fig. 7a*, the shaded region is the trajectory of the root mean square (RMS) EMG signal of each experiment. The decreased and flattened RMS EMG amplitudes demonstrate decreased muscular activity when the volunteer wears SARE.

The mean peak EMG signals are reduced by 32%±15% and 22%±8.2% with and without loading, and the peak value of LES is decreased. The mean iEMG signals are reduced when the volunteer wears SARE, both with and without load. According to equation 9 and Fig. 7b, the efficiencies of wearing SARE are 24.2%±13.6% and 23.6%±9% with and without loading, and the peak value of LES is decreased. This demonstrates that SARE can assist the LES during repetitive lifting with and without load.

IV. DISCUSSION & CONCLUSION

In this paper, we introduced a new design of a light weight soft exosuit which could provide useful compression forces to assist in stoop lifting. And a novel knitting sensor which demonstrates promising potential as a soft, stretchable sensor for monitoring spinal curvature changes, with implications for applications in biomedical and healthcare fields.

A preliminary EMG experiment was conducted to evaluate the exosuit's capacity to reduce iEMG and peak EMG. The exosuit conforms to the curvature of the human back with nearly infinite DOFs and remains parallel to the LES without impeding human movement. Additionally, real-time monitoring of the back's strain sensor facilitates precise actuator control. Various exosuits have been reported in the literature to reduce peak EMG activity during lifting tasks, with reduction percentages ranging from 6-48% [25]. The SARE has achieved a promising 32% reduction in peak EMG activity for loaded lifting tasks. While [25] focused on rigid exoskeletons, it is more appropriate to compare the SARE with other soft exosuits. For example, a soft textile-based, cable-driven exosuit demonstrated a 16% reduction in peak EMG of back extensor [26], and a TSA-based soft exosuit reported a 21.4–25.2% reduction [12]. Additionally, a study involving a cable-driven continuum soft exosuit reported a 30% reduction in compression and shear force of L5-S1. Overall, the SARE's performance demonstrates a promising prospect for reducing muscular workload during lifting tasks. The next stage of system development will focus on optimizing the MASM s-driven smart textiles through improvements to wire diameter, spring rate, spring materials, silicone tube materials, and high-intensity guide tubes to be capable of holding higher hydraulic pressure. The addition of multiple layers of MASM s on the exosuit is also being considered to generate greater compression force. Ultimately, the goal is for the exosuit to facilitate stoop lifting entirely. One area in which this system can be improved in the future is in the implementation of nonlinear modeling and adaptive control which can significantly enhance the system performance during the lifting process. Adding EMG signals and machine learning techniques to predict the bending motion in order to intelligently assist the user when needed is also recommended. Finally, conducting more user studies would demonstrate the effectiveness and benefit of the system for use in practice.